%% file: main.tex
\documentclass[10pt,twocolumn,letterpaper]{article}

\usepackage{cvpr}

\input{preamble}

\input{paper_macros}

\newif\ifwbenchfigmain
\wbenchfigmaintrue

\makeatletter
\pretocmd{\section}{\par\@nobreakfalse}{}{}
\pretocmd{\subsection}{\par\@nobreakfalse}{}{}
\makeatother

\definecolor{linkblue}{rgb}{0.21,0.49,0.74}
\usepackage[breaklinks,colorlinks,allcolors=linkblue]{hyperref}

\def\paperID{}
\def\confName{}
\def\confYear{}

\title{\papertitle}
\author{
Haiyu Xu$^{1,2,*}$ \quad
Yiyang Chen$^{1,*}$ \quad
Jiayi Zhou$^{1,2}$ \quad
Jiaming Ji$^{1,2}$ \quad
Juntao Dai$^{1,2}$ \\
Yaodong Yang$^{1,\dagger}$ \quad
Boyuan Chen$^{1,2,3,\dagger}$ \\[0.5em]
{\small $^{1}$Peking University \quad $^{2}$Physis \quad $^{3}$Beijing Academy of Artificial Intelligence}\\
{\small $^{*}$ Equal contribution \quad
$^\dagger$ Corresponding authors}
}

\begin{document}
\maketitle

\begin{abstract}
  \input{sections/0_abstract}
\end{abstract}

\input{sections/1_introduction}

\input{sections/2_related_work}

\input{sections/3_benchmark}

\input{sections/4_evaluation}

\input{sections/5_experiments}
\input{sections/6_conclusion}

{
  \small
  \bibliographystyle{ieeenat_fullname}
  \bibliography{references}
}

\end{document}

%% file: preamble.tex
\usepackage{graphicx}
\usepackage{amsmath}
\usepackage{amssymb}
\usepackage{array}
\usepackage{booktabs}
\usepackage{multirow}
\usepackage{xspace}
\usepackage{xcolor}

%% file: paper_macros.tex
\newcommand{\benchmarkname}{AxE Bench\xspace}
\newcommand{\papertitle}{AxE: Benchmarking Agent–Environment Generalization of Video World Models via Grounded Events}

\newcommand{\wbenchplus}{WBench-plus\xspace}

\definecolor{agentyellow}{HTML}{FFF200}
\definecolor{envcyan}{HTML}{00E5E5}
\definecolor{jointorange}{HTML}{FFB000}

\newcommand{\agenthl}[1]{%
  \begingroup
  \setlength{\fboxsep}{1pt}%
  \colorbox{agentyellow}{#1}%
  \endgroup
}

\newcommand{\envhl}[1]{%
  \begingroup
  \setlength{\fboxsep}{1pt}%
  \colorbox{envcyan}{#1}%
  \endgroup
}

\newcommand{\jointhl}[1]{%
  \begingroup
  \setlength{\fboxsep}{1pt}%
  \colorbox{jointorange}{#1}%
  \endgroup
}

%% file: sections/0_abstract.tex
Edge cases are not the long tail; they are the norm. An embodied system deployed across changing environments, bodies, objects, and tasks will routinely encounter combinations that were absent from its training data. General-purpose physical intelligence therefore requires world models that reuse prior knowledge, predict consequences, and adapt with little or no additional training. Yet current evaluations of conditional video generation emphasize visual quality, prompt alignment, or isolated physical plausibility, leaving systematic generalization difficult to measure. We introduce \benchmarkname, a benchmark that organizes condition changes along eight factors of the acting agent and of its environment, and pairs every original condition with controlled variants that change one side or both. WBench-plus applies the same structure to an existing multi-turn benchmark, pairing each WBench scenario with variants on the environment side, through its world setting or camera behavior, or on the agent side, through its interaction instructions. Additionally, to score generated futures, we develop an evaluation harness that combines vision-language reasoning with targeted computer-vision evidence, converting each video into temporally localized events and comparing generated events with reference events in both time and semantics. We summarize each model with the AxE Score, event Fidelity weighted by its Stability across original–variant pairs. Across twenty models, generated events match reference events far more often in timing than in content: no model exceeds a Fidelity of 13.01, the highest Stability is 0.55, and the highest AxE Score is 7.10.
Together, these components turn world-model generalization from an informal aspiration into a falsifiable evaluation program for real and simulated interactions. Our project website can be found at: \url{AxEBench.github.io}

%% file: sections/1_introduction.tex
\section{Introduction}
\label{sec:introduction}

\begin{figure}[t]
  \centering
  \includegraphics[width=0.95\linewidth]{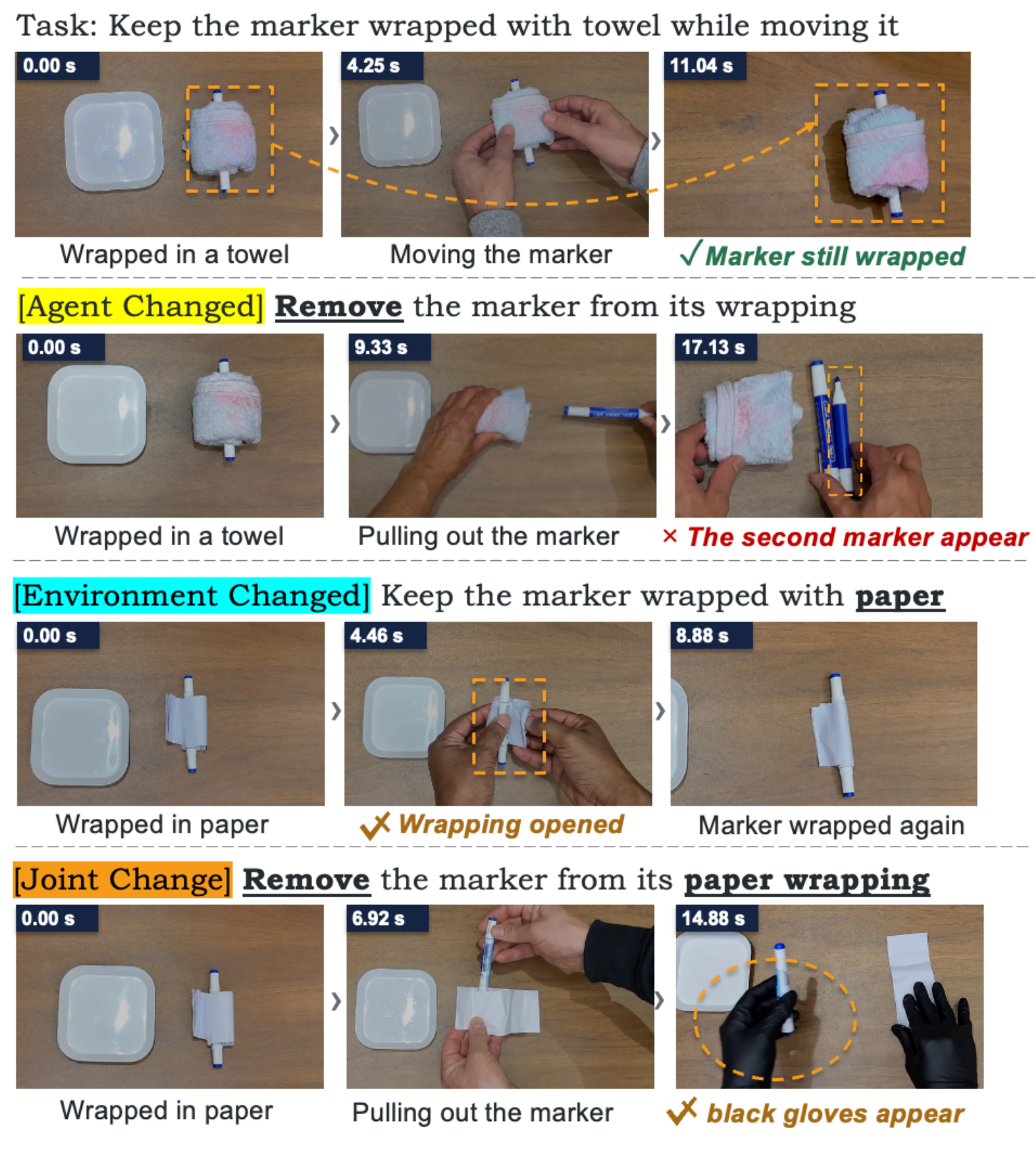}
  \caption{\textbf{Success in one condition does not guarantee generalization.}
Each row shows three generated frames.
The original task is to move a marker while keeping it wrapped
in a towel (top).
Variants change the action to removing the marker
(\agenthl{\textbf{Agent Changed}}),
replace the towel with paper
(\envhl{\textbf{Environment Changed}}),
or combine both changes
(\jointhl{\textbf{Joint Change}}).
Despite success in the original condition, the variants
exhibit object duplication, unintended unwrapping,
and inconsistent agent appearance, respectively.}
  \label{fig:teaser}
\end{figure}

\begin{figure*}[t]
  \centering
  \includegraphics[width=\linewidth]{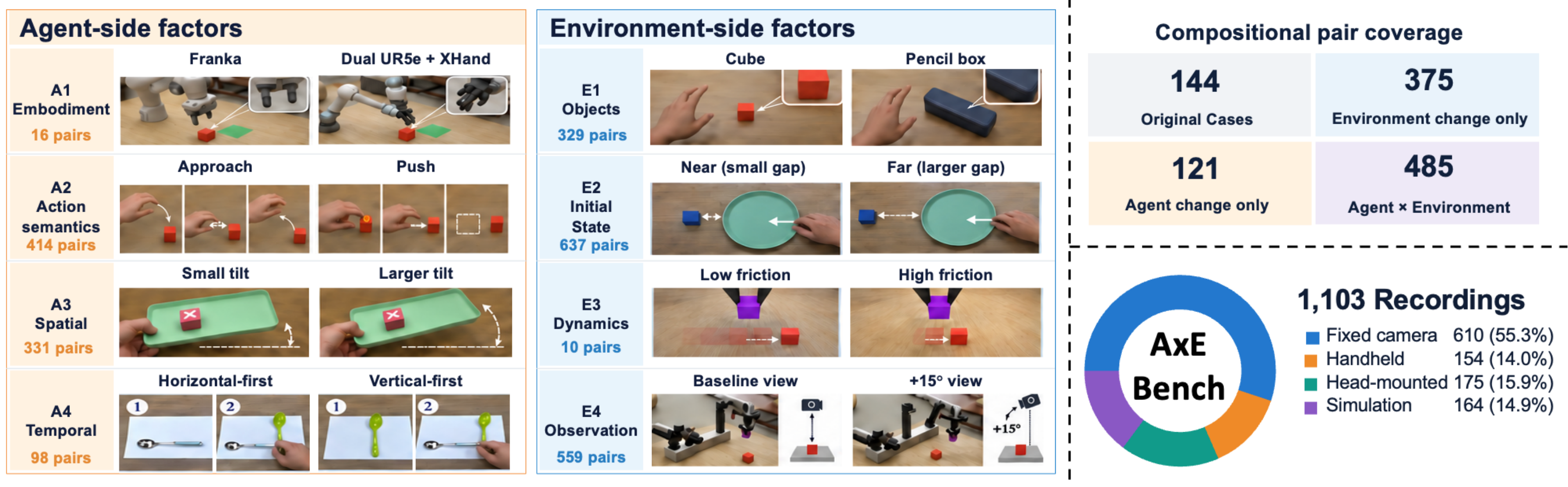}
  \caption{\textbf{Data taxonomy and coverage.}
  We organize changes relative to an original condition into eight agent-side
  and environment-side factor families; each is shown with an original--variant 
  pair and the number of pairs that change it. The presence of changes on
  either side defines four regimes: original, agent generalization,
  environment generalization, and agent--environment composition. The ring breaks the recordings down by
  capture setup.}
  \label{fig:taxonomy}
\end{figure*}

Anticipating the consequences of actions is a central goal on the path toward
artificial general intelligence, and video generation offers a concrete way to
pursue it: given an observation and a description of an action, a model
generates a possible continuation of the scene. Since video generation models
were first explored as world simulators~\cite{openai2024sora}, later systems
have advanced physical fidelity~\cite{openai2025sora2}, action-conditioned
robot simulation~\cite{gao2026dreamdojo}, and interactive
generation~\cite{genie3,genie4}. These advances bring video generation closer
to embodied intelligence and open-world interaction, where predictions must
hold across diverse operating conditions.

Such applications make generalization a central requirement.
As illustrated in Fig.~\ref{fig:teaser}, success in an original
condition can coexist with failures in object consistency,
action execution, and agent appearance when the action or environment change.
The same interaction may involve a different embodiment, an unfamiliar object, a changed physical setting, or a new viewpoint. This raises a central question for video-based world simulation: how reliably do models generate the appropriate consequences of an interaction when its conditions change, individually or in combination?

Answering this question requires evaluating performance across
explicitly related conditions. Existing benchmarks provide important foundations: VBench and EvalCrafter assess video quality and prompt alignment ~\cite{huang2023vbench,liu2023evalcrafter}, while VideoPhy,
Physics-IQ, and WorldModelBench examine physical plausibility and world-modeling capabilities
~\cite{bansal2024videophy,motamed2025physicsiq,li2025worldmodelbench}. Our focus is on how these capabilities generalize under controlled changes. Evaluating diverse scenarios alone does not isolate which changes affect performance or how multiple changes interact. We therefore organize evaluation around explicit variations of reference conditions, enabling comparisons of baseline competence, sensitivity to individual factors, and robustness to their composition.

We introduce \benchmarkname{} (Agent × Environment), a benchmark for evaluating the
generalization of world-simulation capabilities in conditional video generation. 
Each condition
is described by eight factors, four of the acting agent and four of the
environment and how it is perceived, and each original condition is paired
with variants that change the agent side, the environment side, or both.

We further apply this framework to extend an existing benchmark to generalization evaluation. Specifically, we construct \wbenchplus{} by augmenting each multi-turn scenario in WBench~\cite{ying2026wbench} with variations in world settings and camera behavior on the environment side, and interaction instructions on the agent side.

Evaluating these changes also requires evidence beyond overall visual impressions. A decisive error may occur during a brief contact, emerge after an object becomes occluded, or accumulate across a sequence of events. We therefore develop an evaluation harness that combines computer-vision evidence with vision-language reasoning to assess generated entities, interactions, and event progression. Our contributions are threefold:
\begin{itemize}
    \item \textbf{A benchmark for generalization in video-based world simulation.}
    We introduce \benchmarkname{}, which structures evaluation conditions into eight atomic factors spanning the agent and environment.
    As part of the benchmark, we construct \wbenchplus{} (Sec.~\ref{sec:wbench-plus}), an extension of WBench with paired original and modified multi-turn tasks.

    \item \textbf{An evidence-based evaluation harness.}
    We introduce WM-PerceptHarness, which combines vision-language reasoning with targeted visual evidence gathering to assess the temporal and semantic fidelity of generated events, with traceable evidence for diagnosis. On top of it, the AxE Score combines event Fidelity with its Stability across original–variant pairs into a single measure of generalization.

    \item \textbf{A systematic analysis of model generalization.}
    Across twenty models, generated events often fall at the right time but
    rarely show the right content: Timing is 38.8 to 52.9 for the text-and-image-to-video models, while no model exceeds a Fidelity of 13.01. The scores also do not hold under a change: the highest Stability is 0.55, and the highest AxE Score is 7.10.

\end{itemize}

%% file: sections/2_related_work.tex
\begingroup
\setlength{\parskip}{0pt}
\makeatletter
\def\cvprsection{\@startsection{section}{1}{\z@}
  {-10pt}{7pt}{\large\bf}}
\def\cvprsubsection{\@startsection{subsection}{2}{\z@}
  {-8pt}{5pt}{\elvbf}}
\makeatother

\section{Related Work}
\label{sec:related-work}

\paragraph{Video Generation Models and World Models}

A world model is a learned predictive model of an environment that
captures its dynamics and predicts future observations or latent
states \citep{hafner2020dreamer}.
Through multi-step prediction and imagined rollouts, world models
support control and policy learning
\citep{hafner2020dreamer},
policy evaluation \citep{li2025worldeval},
and environment simulation and interactive generation
\citep{bruce2024genie}.
Video generation models have become a common way to build such models:
they predict future frames from an image and a text
prompt~\citep{wan2025wan,hunyuan2025video15,hacohen2026ltx2} and have been
proposed as general-purpose simulators of the physical
world~\citep{openai2024sora,nvidia2025cosmospredict25}, and additional
conditioning on actions or camera motion makes the generated world
interactive~\citep{genie3,he2026matrixgame2,robbyant2026lingbotworld}.
These applications require reliable generalization across world
settings, interaction demands, and observation conditions, motivating
our systematic evaluation under controlled shifts.

\paragraph{World-Model Evaluation}

\noindent\textit{Evaluation of model capabilities:}
WorldModelBench~\citep{li2025worldmodelbench} evaluates instruction
following, physical adherence, and common sense.
WorldScore~\citep{duan2025worldscore} assesses controllability, visual
quality, and motion dynamics across video and 3D/4D world generation.
WBench~\citep{ying2026wbench} evaluates interactive video world models
on multi-turn tasks spanning navigation, subject actions, event edits,
and perspective switches. \textit{Evaluation of downstream applications:}
WorldSimBench~\citep{qin2025worldsimbench} combines perceptual
evaluation with closed-loop embodied task evaluation.
WorldArena~\citep{shang2026worldarena} jointly evaluates video quality
and functional utility in robotic manipulation. \textit{Generalization and model generality:}
\citet{kang2025physical} investigate physical prediction in video
diffusion models under in-distribution, out-of-distribution, and
combinatorial generalization settings.
WorldTest~\citep{warrier2026worldtest} proposes a protocol for evaluating
whether agents can use environmental knowledge acquired through
reward-free interaction to solve diverse environment-level queries.

\par
\endgroup

%% file: sections/3_benchmark.tex
\section{The AxE Bench}
\label{sec:benchmark}

\begin{figure*}[t]
  \centering
  \includegraphics[width=\linewidth]{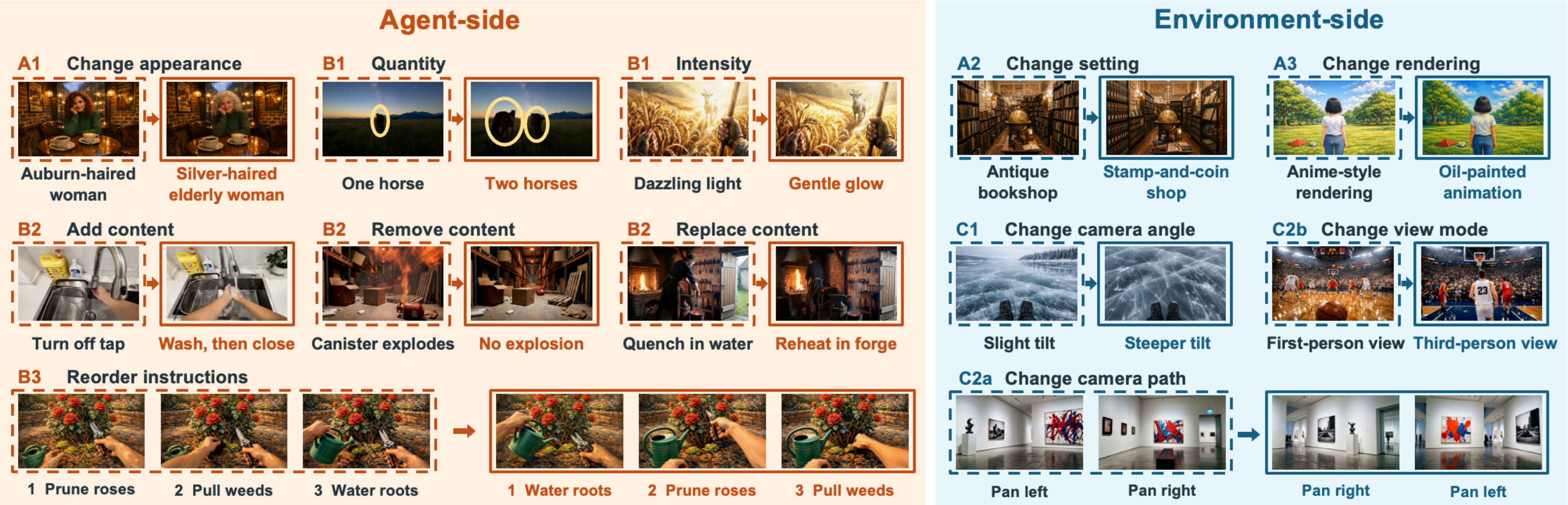}
  \caption{\textbf{Construction of \wbenchplus.}
  We show representative original–variant pairs selected from the newly constructed \wbenchplus cases. The agent-side examples (left) illustrate Change subject appearance (A1), Quantity and Intensity (B1), Add, Remove, and Replace content (B2), and Reorder instructions (B3). The environment-side examples (right) illustrate Change setting (A2), Change rendering (A3), Change camera angle (C1), Change camera path (C2a), and Change view mode (C2b). Dashed outlines mark the original WBench~\cite{ying2026wbench} cases or instruction sequences; solid outlines mark their \wbenchplus variants.}
  \label{fig:wbench-plus}
\end{figure*}

\subsection{Agent--Environment Generalization Space}
\label{sec:generalization-space}

\paragraph{Interaction and model.}
A condition is an interaction between an agent and its environment, seen
through a camera: at step $t$ the camera gives an observation $o_t$ of the
environment state, and the agent issues an action $a_t$ that changes it.
A video world model
$G_\theta$ receives the first observation and the action sequence and generates
a video:
\begin{equation}
\hat{V} = G_\theta\big(o_1, \mathbf{a}_{1:T}\big)
        = (o_1, \hat{o}_2, \dots, \hat{o}_T)
\label{eq:generation}
\end{equation}

\paragraph{Factors.}
What happens in a condition is determined by two things: the actions that the
agent issues, and the environment in which they take effect, that is, the
initial state, how the state responds to an action, and how it is observed. A
variant can therefore differ from its original condition on the agent side, in
the action sequence, or on the environment side, in everything else. We divide
each side into four factors (Fig.~\ref{fig:taxonomy}). The four agent-side
factors change the action sequence: who issues the actions (\textbf{Embodiment}), what they are
and what goal they serve (\textbf{Action Semantics}), how an action with the same goal is executed, such as its amplitude or direction (\textbf{Spatial}), and how the steps are ordered or repeated (\textbf{Temporal}). The four environment-side factors leave the actions as they are and change the rest: which task-relevant objects the first state contains (\textbf{Objects}), how they are arranged (\textbf{Initial State}), the physical conditions under which the state responds to an action, such as friction or mass (\textbf{Dynamics}), and the viewpoint and motion of the camera through which the state is observed (\textbf{Observation}). 

\subsection{Data Collection and Pairing}
\label{sec:data}

\paragraph{Recordings.}
The benchmark contains 1,103 recordings. Real recordings show people
manipulating everyday objects and are captured with a fixed camera (610), a
handheld moving camera (154), or a head-mounted camera (175). The remaining 164
are simulated robot manipulation, which lets us vary what real capture cannot
control cleanly: the robot arm, the friction, and the camera viewpoint, which
is rotated by a known angle.

\paragraph{Pairs and evaluation settings.}
Recordings are organized in task groups, in which one recording is the original
condition and the others are variants that keep the task and change designated
factors. This yields 981 original--variant pairs over 144 original conditions.
A pair that changes several factors is counted under each of them
(Fig.~\ref{fig:taxonomy}): initial state (637 pairs), observation (559), and
action semantics (414) are the most frequent, while embodiment (16) and
dynamics (10) are the rarest. We evaluate two input settings on the same
recordings and pairs. In the text-and-image-to-video setting, a model receives
the first frame and the text prompt; in the camera-conditioned setting, it
additionally receives a camera trajectory, which is the recorded trajectory for
the 329 moving-camera recordings and a static one for the 774 fixed-camera
recordings, real and simulated.

\paragraph{Extension to multi-turn interaction.}
\label{sec:wbench-plus}
To verify that the original--variant design is not tied to our own recordings,
we apply it to WBench~\cite{ying2026wbench}, a multi-turn benchmark that was
not built for generalization. Each of its 289 cases becomes an original
condition and is paired with variants that apply one designated edit and keep
the number of turns (Fig.~\ref{fig:wbench-plus}): world setting and view and
control change the environment side, and interaction instructions change the
agent side. The resulting \wbenchplus contains 884 variants (283
world-setting, 283 view-and-control, and 318 instruction variants) that cover
all 289 original cases and comprise 3,273 turns. Together with the 1,058 turns in the original cases, the full suite contains 4,331 turns.

%% file: sections/4_evaluation.tex
\section{Grounded Event Evaluation}
\label{sec:evaluation}

\newcommand{\scoringfigure}{%
\begin{figure*}[t]
  \centering
  \includegraphics[width=0.93\linewidth]{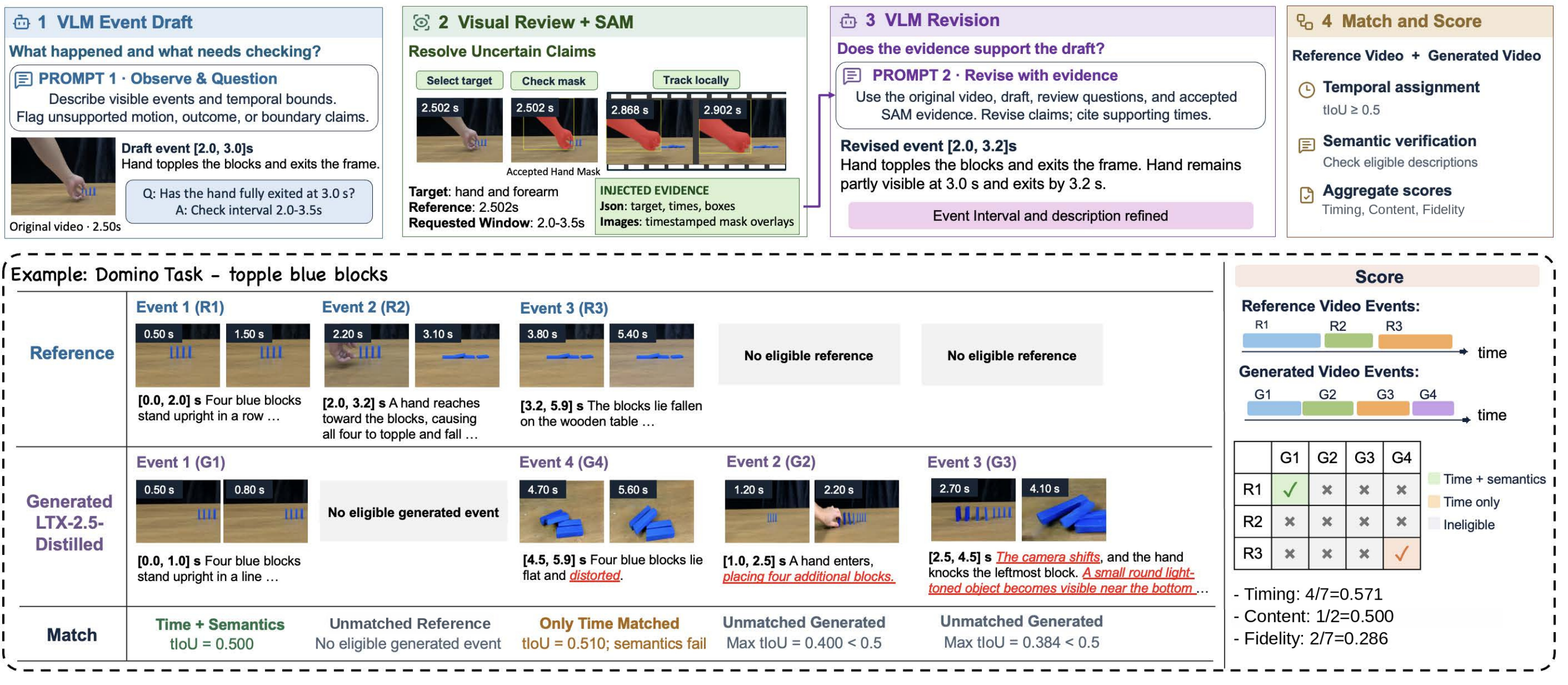}
  \caption{\textbf{Grounded event evaluation.}
  A vision-language model converts each video into timestamped event
  descriptions in three steps: it drafts the events and flags claims that need
  checking, gathers question-driven SAM evidence (target selection, mask check,
  local tracking) for those claims, and revises the events with the accepted
  evidence. Reference and generated events are then compared. Bottom: a domino task with three reference and four
  generated events, its match matrix, and the resulting scores.}
  \label{fig:scoring}
\end{figure*}}
\scoringfigure

A generated video can look plausible and still get the interaction wrong, and
a single judgment of the whole clip cannot say where. Our harness,
WM-PerceptHarness, therefore converts every video into a timeline of events
supported by visual evidence and compares the events of a generated video with
those of its reference (Fig.~\ref{fig:scoring}).

\subsection{Evidence-Grounded Event Annotation}
\label{sec:annotation}

An event is a triple of start time, end time, and a text description of what is
observed in that interval, with times on the clock of the original video.
Annotation proceeds in three steps. First, a vision-language model (VLM)
watches the video, drafts a timeline, and lists the claims in its draft that
the frames it has seen do not settle, such as whether a hand has fully left the
frame or where an event ends; each becomes a review question with a time
window. Second, for each question the harness gathers evidence with a
segmentation model (SAM~3.1~\cite{meta2026sam31}): it selects the target named
in the question at a reference frame, checks the mask, and tracks the target
over the requested window, which yields timestamped frames with mask overlays.
The segmentation model does not decide what an action means or whether it
succeeded. Third, the VLM receives the video, its draft, the questions, and the
accepted evidence, and revises event descriptions and boundaries, citing the
times that support each change. The same procedure annotates reference
recordings and generated videos.

\subsection{Event Matching and Scores}
\label{sec:matching}

Let a reference timeline have $N_r$ events and a generated timeline $N_g$
events. For every pair of a reference event and a generated event we compute
the temporal intersection over union (tIoU) of their intervals; pairs with
$\mathrm{tIoU} \geq 0.5$ are temporally eligible. A semantic judge then reads
the two descriptions of each eligible pair, 
and decides whether they report the same object, action, direction, and outcome.

We compute two assignments.
The \emph{temporal} assignment uses
all eligible pairs and matches $M_t$ of them, of which $S_t$ also pass the
judge. The \emph{joint} assignment is computed separately, using only pairs
that are eligible and pass the semantic judge, and matches $M_j$ pairs. The scores are
\begin{equation}
\begin{aligned}
\mathrm{Timing} &= \frac{2 M_t}{N_r + N_g}, \qquad
\mathrm{Content} = \frac{S_t}{M_t}, \\[2pt]
\mathrm{Fidelity} &= \frac{2 M_j}{N_r + N_g}.
\end{aligned}
\end{equation}

Timing is an F1 score over time-matched events: it measures whether the
generated video has events where the reference has them, regardless of
what they show. Content is the share of time-matched events whose
descriptions also agree. Fidelity is an F1 score over events matched in
both time and content. All
three are reported $\times 100$.

\paragraph{Paired comparison.}
To measure the effect of a change, each variant is compared only with its own
original condition. Let $f(x)$ be the Fidelity of a single video $x$, and let
$(v_i, o_i)$, $i = 1, \dots, n$, be the pairs of a variant $v_i$ and its
original condition $o_i$. The \emph{Stability} of a model is
\begin{equation}
\label{eq:stability}
\mathrm{Stability}
  = 1 - \frac{\sum_{i} \bigl| f(v_i) - f(o_i) \bigr|}
             {\sum_{i} \bigl( f(v_i) + f(o_i) \bigr)},
\end{equation}
the share of the paired score that is reached on both sides. It is one
when no score changes within a pair and zero when every pair scores on one
side only, whichever side that is. The \emph{AxE Score} is the product of the two,
\begin{equation}
\text{AxE Score} = \text{Fidelity} \times \text{Stability},
\label{eq:axe}
\end{equation}
the score of a model weighted by how well it holds under a change.

\paragraph{Complementary metrics.}
Event scores do not measure visual quality, so we also report PSNR against the
reference recording, Dynamic Degree and Motion
Smoothness~\cite{huang2023vbench}, Semantic Adherence and Physical Commonsense
on a 1--5 scale~\cite{bansal2026videophy2}, the camera motion error
CamMC~\cite{zheng2024cami2v} for camera-conditioned models, and the GeCo
structure error~\cite{gu2026geco} relative to the real recording of the same
task.

\subsection{Agreement with Human Judgments}
\label{sec:human-agreement}

We validate the semantic judge and Fidelity against blinded human annotations (1,360 annotations across 1,060 items).

\noindent\textbf{Semantic judge.} On 578 event pairs, the judge achieves 87\% agreement with human judgments (Cohen’s $\kappa = 0.72$).

\noindent\textbf{Fidelity.} Human annotators compare two generated videos against a reference, selecting the closer one or a tie. Across nine models, rankings by Fidelity strongly correlate with human win-rate rankings (Spearman $\rho = 0.90$).

%% file: sections/5_experiments.tex
\newcommand{\casefigure}{%
\begin{figure*}[t]
  \centering
  \includegraphics[width=\linewidth]{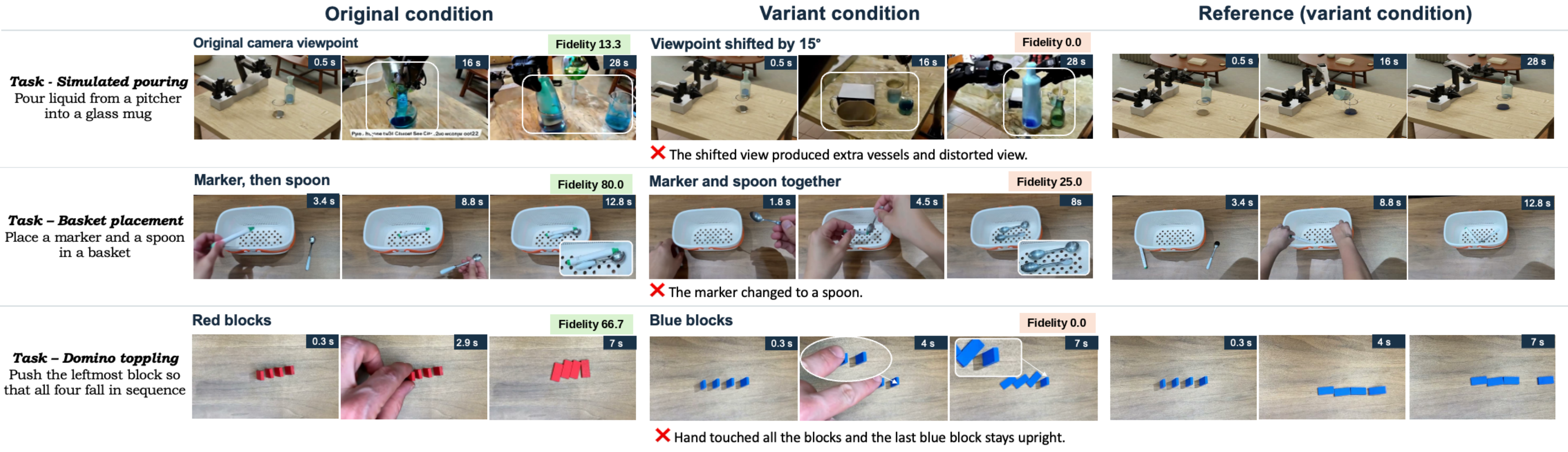}
  \caption{\textbf{Typical examples.} Each row shows key frames of the generated video for the original condition (left) and the variant condition (middle), with their Fidelity, and the reference recording of the variant condition (right). Changed factor by row: observation (simulated camera rotated by 15°; LTX-2.5), action semantics (marker and spoon placed together instead of in sequence; LTX-2.5), initial state (block color; MiniMax H3).}
  \label{fig:cases}
\end{figure*}}

\section{Experiments}
\label{sec:experiments}

\subsection{Models and Evaluation Setup}
\label{sec:setup}

\input{tables/results_main}

We evaluate twenty video generation models on our recorded benchmark. Thirteen take a first frame and
a text prompt: nine open models, MiniMax H3~\cite{minimax2026h3}, LTX-2.5
(distilled)~\cite{hacohen2026ltx2},
VPP2~\cite{guo2026vpp2}, Cosmos~3 Super and Nano~\cite{nvidia2026cosmos3}, Wan2.2
TI2V-5B~\cite{wan2025wan}, HunyuanVideo~1.5~\cite{hunyuan2025video15},
Cosmos-Predict2.5-2B~\cite{nvidia2025cosmospredict25}, and
GigaWorld-0~\cite{gigaworld2025gigaworld0}, and four commercial models, Wan~3.0~\cite{alibaba2026wan3}, Kling~4.0~\cite{kling2026kling4}, Seedance~2.5~\cite{bytedance2026seedance25}, and HappyHorse~1.1~\cite{alibaba2026happyhorse11}. Seven
further models additionally take a camera trajectory: AlayaWorld~\cite{alayaworld2026},
Matrix-Game-2.0~\cite{he2026matrixgame2}, Wan2.2 Fun
Camera~\cite{alibabapai2025wan22fun}, SANA-WM~\cite{zhu2026sanawm},
LingBot-World~\cite{robbyant2026lingbotworld}, Evoke~\cite{yin2026evoke}, and
Evoke Turbo~\cite{alayalab2026evoketurbo}.

The vision-language model that
drafts and revises the events and the semantic judge are both Doubao-Seed-2.1-Pro~\cite{bytedance2026doubaoseed21pro};
visual evidence comes from SAM~3.1~\cite{meta2026sam31}. The judge runs at temperature zero.

\subsection{Main Results}
\label{sec:main-results}

\ifwbenchfigmain\casefigure\fi

\paragraph{Reproducing the reference events is hard for every model.}
The four commercial models score highest, with a Fidelity of 12.22 to 13.01 (Table~\ref{tab:main-results}). Among the open models, the best is 10.06
(MiniMax H3), the second 5.59 (LTX-2.5), and the other seven are below 3.0.
Models that additionally take a camera trajectory do not score higher: the
best reaches 4.45 (AlayaWorld), below the two best open models, and Evoke
Turbo has both the lowest camera motion error (0.850) and the lowest Fidelity
(0.68). Timing lies between 38.8 and 52.9 for the thirteen
text-and-image-to-video models, but Content is at most 26.2:
generated events fall at roughly the times of the reference events, and their
content usually differs.
Once Stability is taken into account, the gap widens further: the highest AxE Score is 7.10 (Wan 3.0), and Sec.~\ref{sec:generalization-results} examines why.
In Fig.~\ref{fig:cases}, a marker turns into a second
spoon, a rotated view gains vessels that were never in the scene, and a hand
touches every block instead of pushing the first one.

\begin{figure}[t]
  \centering
  \includegraphics[width=0.9\linewidth]{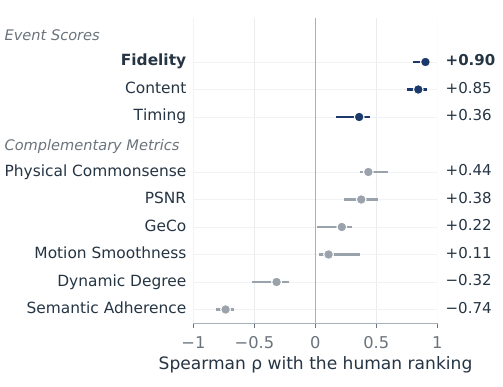}
  \caption{\textbf{Agreement of each metric with human preference.} Spearman $\rho$ between the ranking of the nine open text-and-image-to-video models by each metric and their ranking by human win rate (Sec.~\ref{sec:human-agreement}). Bars are 95\% intervals from resampling the human comparisons. GeCo is counted as lower-is-better, all other metrics as higher-is-better.}
  \label{fig:metric-human}
\end{figure}

\paragraph{Complementary metrics separate the models less and do not follow human preference.}
These metrics describe how a video looks and moves, not whether it shows the
instructed interaction. Some barely separate the models: across the thirteen
text-and-image-to-video models, Motion Smoothness lies between 0.978 and 0.996
and Semantic Adherence between 2.93 and 3.01, while Fidelity ranges from 1.48
to 13.01. Where they do rank the models, the ranking does not match human preference
(Fig.~\ref{fig:metric-human}): Fidelity follows the human ranking
($\rho = 0.90$), but no complementary metric exceeds 0.44, and Semantic Adherence
runs against it. PSNR and GeCo favor videos that move little: among the text-and-image-to-video models, the three with the highest PSNR are among the four with the lowest Dynamic Degree.

\subsection{Generalization under Paired Changes}
\label{sec:generalization-results}
\ifwbenchfigmain\else\casefigure\fi

We compare each variant with its own original condition as in Sec.~\ref{sec:matching}, on the same 981 pairs for all models. 

\begin{figure*}[t]
  \centering
  \includegraphics[width=0.9\linewidth]{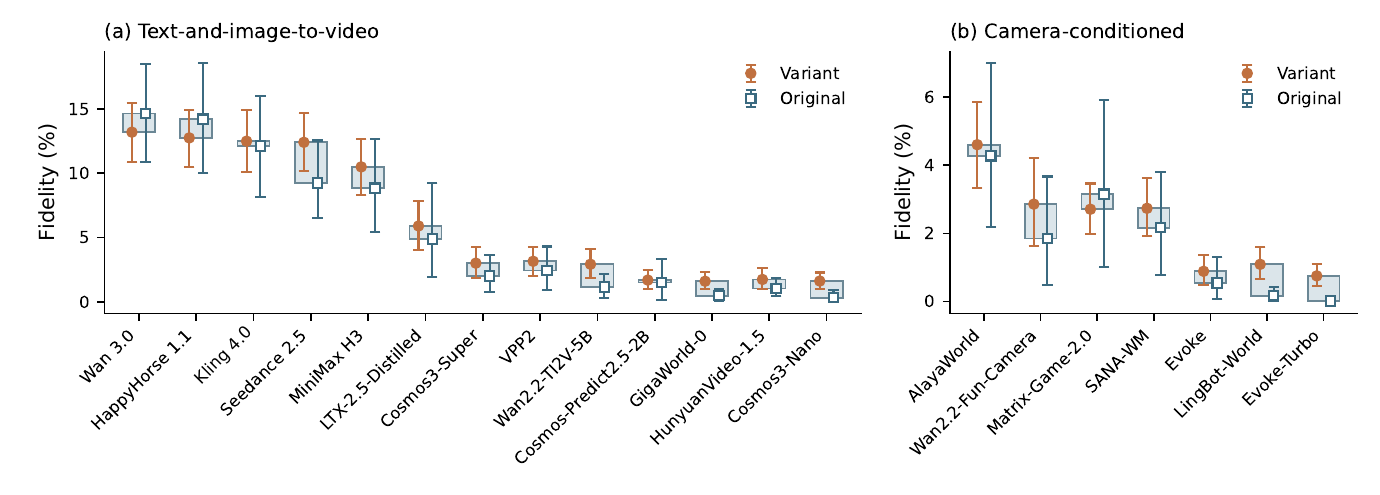}
  \vspace{-0.45em}
  \caption{\textbf{Fidelity on the original and on the variant conditions.} For each model, the two markers give the Fidelity over all original videos of the 981 pairs and over all their variant videos, each computed from pooled event counts; bars are 95\% intervals from resampling task families. Models are ordered by their mean Fidelity over the pairs. A model that generalizes scores high on both sides with a small difference between them; a small difference at a low score does not indicate this. (a) Text-and-image-to-video; (b) camera-conditioned, on a different axis range.}
  \label{fig:pair-outcomes}
\end{figure*}

\paragraph{Pooled scores separate three kinds of behavior.} Figure~\ref{fig:pair-outcomes} gives, for each model, the Fidelity over all original videos of the 981 pairs and over all their variant videos. We read the two scores together. A model with high Fidelity on both sides and a small gap between them generates the instructed events and keeps doing so when the condition changes. A model with high Fidelity but a large gap is capable but not robust: Seedance 2.5 has the largest gap of all twenty models (3.2 points), although its Fidelity is within one point of the other commercial models, and it also has the lowest stability of the four (0.42, Table~\ref{tab:main-results}). A model with low Fidelity on both sides lacks the ability in the first place, and its small gap is not evidence of robustness: for the models at the bottom of Fig.~\ref{fig:pair-outcomes}, both scores are close to zero. The gap is therefore informative only together with the level, which is what the AxE Score combines.

\begin{figure}[!t]
  \centering
  \includegraphics[width=0.8\linewidth]{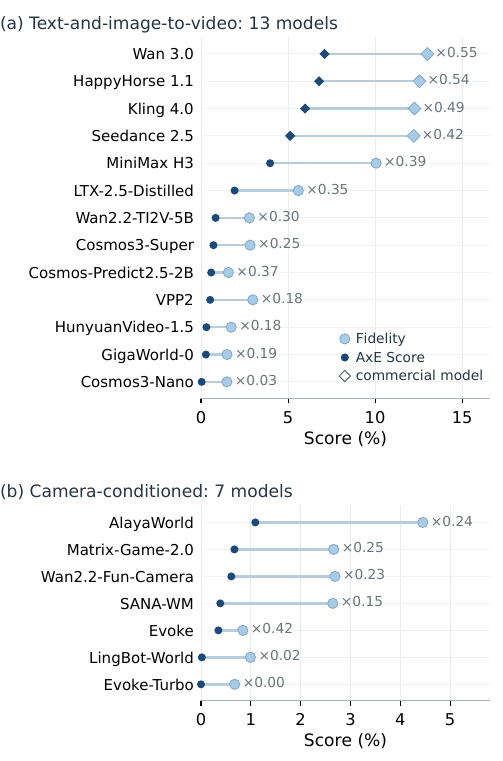}
  \caption{\textbf{Fidelity and AxE Score.}
   Light markers: Fidelity (Table~\ref{tab:main-results}). Dark markers: the AxE Score, Fidelity times Stability (Eq.~\eqref{eq:stability}); the Stability value is printed beside each row. Models are ordered by AxE Score; diamonds mark the commercial models. For scores close to zero, Stability rests on few pairs. The axis runs from 0 to 15 in (a) and from 0 to 5 in (b).}
  \label{fig:score-vs-gap}
\end{figure}

\paragraph{Within pairs, even the strongest models hold about half of their score.}
Stability (Eq.~\eqref{eq:stability}) measures whether a model reproduces both conditions of
the same pair, and
Fig.~\ref{fig:score-vs-gap} multiplies it onto Fidelity. It ranges
from 0.39 (MiniMax H3) to 0.55 (Wan~3.0) for the five highest models and is
0.03 for Cosmos~3 Nano. The AxE Score of Wan~3.0 is 7.10 against a Fidelity of 13.01. Among the text-and-image-to-video models, the six highest models keep their order; below them, VPP2 falls
from seventh to tenth and Wan2.2 TI2V-5B rises from ninth to seventh. Among
the camera-conditioned models, the highest AxE Score is 1.09 (AlayaWorld). 
For models that score on few pairs, Stability is unreliable
(0.42 for Evoke), and we do not interpret it.

\subsection{Multi-Turn Generalization on \wbenchplus}
\label{sec:wbench-plus-results}
\input{tables/wbench_plus_leaderboard}

For the multi-turn evaluation, Table~\ref{tab:wbench-events-paired} reports
20 models from the main benchmark together with two additional models,
EchoWM and its four-step autoregressive variant,
EchoWM Flash~\cite{zhang2026echowm}. In total, we report 22 models on
\wbenchplus,
expanding our generalization design and event evaluation.

\paragraph{Event evaluation extends to tasks without ground-truth videos.}
\wbenchplus also demonstrates how the event scores can be applied when
ground-truth videos are unavailable. Each condition's turn instructions
specify the required behavior, and its turn windows provide the reference
intervals. Generated videos are converted into event timelines and evaluated
against these instruction-derived references using the temporal and semantic
matching of Sec.~\ref{sec:matching}. 
Timing measures alignment with the
reference intervals, Content assesses whether time-matched events
fulfill the instructions, and Fidelity requires both. 
This provides a common
evaluation structure for text-and-image-to-video and camera-conditioned
models, while allowing each modified task to be scored against its own
requirements.

\paragraph{The generalization design reveals sensitivity beyond our own recordings.}
Applying the agent--environment variation design to WBench reveals substantial
sensitivity in an existing multi-turn benchmark. Across changes to world
settings, camera behavior, and interaction instructions, the highest Stability
is only 0.62 among the commercial models, 0.53 among the open
text-and-image-to-video models, and 0.67 among the camera-conditioned models.
Even the most stable models therefore show substantial differences between
paired conditions. These results show that the original--variant design exposes
limitations that evaluation on individual conditions alone would leave
unmeasured, extending the analysis of generalization from our recorded
interactions to WBench's multi-turn tasks.


%% file: tables/results_main.tex
\begin{table*}[t]
\centering
\footnotesize
\caption{
Main results on the 1,103 recordings. Event scores (Sec.~\ref{sec:matching}, $\times 100$) are pooled over all recordings; Stability (Eq.~\eqref{eq:stability}) is computed on the 981 pairs.
The complementary metrics are averaged over the same recordings. MS: Motion Smoothness; SA: Semantic Adherence; PC: Physical Commonsense; DD: Dynamic Degree. CamMC is computed on the 329 moving-camera recordings only. Rows are sorted by Fidelity; the first four text-and-image-to-video models are commercial. The best value of each column within each of the two groups is in bold.}
\label{tab:main-results}
\setlength{\tabcolsep}{3pt}
\resizebox{\linewidth}{!}{%
\begin{tabular}{@{}l ccc c cc c ccccc cc@{}}
\toprule
 & \multicolumn{3}{c}{Event scores} & & \multicolumn{2}{c}{Generalization} & & \multicolumn{7}{c}{Complementary metrics} \\
\cmidrule(lr){2-4}\cmidrule(lr){6-7}\cmidrule(l){9-15}
Model & Fidelity $\uparrow$ & Timing $\uparrow$ & Content $\uparrow$ & & Stability $\uparrow$ & AxE Score $\uparrow$ & & PSNR $\uparrow$ & MS $\uparrow$ & SA $\uparrow$ & PC $\uparrow$ & CamMC $\downarrow$ & DD & GeCo \\
\midrule
\multicolumn{15}{@{}l}{\textit{Text-and-image-to-video}} \\
Wan 3.0 & \textbf{13.01} & 49.71 & \textbf{26.16} &  & \textbf{0.55} & \textbf{7.10} &  & 17.17 & 0.995 & 2.94 & \textbf{4.04} & -- & 0.637 & 1.043 \\
HappyHorse 1.1 & 12.55 & \textbf{52.92} & 23.72 &  & 0.54 & 6.79 &  & 15.44 & 0.990 & 2.94 & 3.97 & -- & 0.792 & 1.411 \\
Kling 4.0 & 12.26 & 49.92 & 24.56 &  & 0.49 & 5.98 &  & \textbf{18.08} & \textbf{0.996} & 2.93 & 3.99 & -- & 0.545 & 0.790 \\
Seedance 2.5 & 12.22 & 52.26 & 23.38 &  & 0.42 & 5.12 &  & 17.06 & \textbf{0.996} & 2.93 & 4.01 & -- & 0.579 & 0.994 \\
\cmidrule{1-15}
MiniMax H3 & 10.06 & 52.04 & 19.30 &  & 0.39 & 3.97 &  & 15.86 & 0.994 & 2.94 & 3.97 & -- & 0.652 & 1.069 \\
LTX-2.5-Distilled & 5.59 & 41.06 & 13.61 &  & 0.35 & 1.93 &  & 15.67 & 0.992 & 2.96 & 3.86 & -- & 0.762 & 1.411 \\
VPP2 & 2.98 & 45.02 & 6.57 &  & 0.18 & 0.53 &  & 14.84 & 0.978 & 2.99 & 3.50 & -- & 0.931 & 1.179 \\
Cosmos3-Super & 2.82 & 43.31 & 6.52 &  & 0.25 & 0.72 &  & 15.41 & 0.992 & 2.99 & 3.66 & -- & 0.896 & 1.748 \\
Wan2.2-TI2V-5B & 2.78 & 38.75 & 7.18 &  & 0.30 & 0.84 &  & 15.87 & 0.995 & 2.96 & 3.85 & -- & 0.581 & 1.280 \\
HunyuanVideo-1.5 & 1.73 & 42.39 & 4.09 &  & 0.18 & 0.31 &  & 14.91 & 0.995 & 2.97 & 3.82 & -- & 0.821 & 0.756 \\
Cosmos-Predict2.5-2B & 1.57 & 44.75 & 3.52 &  & 0.37 & 0.59 &  & 13.21 & 0.993 & 3.00 & 3.82 & -- & 0.844 & 1.691 \\
GigaWorld-0 & 1.49 & 40.95 & 3.61 &  & 0.19 & 0.28 &  & 16.01 & 0.991 & \textbf{3.01} & 3.46 & -- & 0.855 & 1.013 \\
Cosmos3-Nano & 1.48 & 42.19 & 3.52 &  & 0.03 & 0.04 &  & 14.30 & 0.990 & 3.00 & 3.61 & -- & 0.937 & 1.885 \\
\midrule
\multicolumn{15}{@{}l}{\textit{Camera-conditioned image-to-video}} \\
AlayaWorld & \textbf{4.45} & 32.32 & \textbf{13.78} &  & 0.24 & \textbf{1.09} &  & \textbf{17.64} & \textbf{0.996} & 2.62 & 4.26 & 1.116 & 0.240 & 0.380 \\
Wan2.2-Fun-Camera & 2.69 & \textbf{46.05} & 5.81 &  & 0.23 & 0.61 &  & 13.18 & 0.992 & 2.97 & 3.86 & 1.080 & 0.830 & 1.437 \\
Matrix-Game-2.0 & 2.66 & 31.03 & 8.58 &  & 0.25 & 0.67 &  & 15.33 & 0.993 & 2.44 & \textbf{4.45} & 1.151 & 0.290 & 1.401 \\
SANA-WM & 2.65 & 45.88 & 5.73 &  & 0.15 & 0.39 &  & 14.36 & 0.985 & 2.94 & 3.46 & 0.966 & 0.802 & 1.556 \\
LingBot-World & 0.99 & 41.24 & 2.41 &  & 0.02 & 0.02 &  & 13.24 & 0.982 & 2.92 & 3.88 & 1.212 & 0.762 & 2.689 \\
Evoke & 0.84 & 44.57 & 1.89 &  & 0.42 & 0.35 &  & 14.71 & 0.993 & \textbf{3.01} & 3.80 & 0.898 & 0.866 & 0.960 \\
Evoke-Turbo & 0.68 & 42.98 & 1.54 &  & 0.00 & 0.00 &  & 14.73 & 0.993 & 3.00 & 3.76 & \textbf{0.850} & 0.869 & 1.100 \\
\bottomrule
\end{tabular}%
}
\end{table*}

%% file: tables/wbench_plus_leaderboard.tex
\begin{table}[t]
\centering
\caption{
\textbf{Results on the multi-turn \wbenchplus cases.}
Rows are sorted by AxE Score within each of the three groups; the first four text-and-image-to-video models are commercial. The best value of each column within each of the three groups is in bold.
}
\label{tab:wbench-events-paired}
\begingroup
\fontsize{7}{8}\selectfont
\setlength{\tabcolsep}{1pt}
\renewcommand{\arraystretch}{1.1}
\begin{tabular*}{\columnwidth}{@{\extracolsep{\fill}}lrrrrr@{}}
\toprule
Model & Fidelity $\uparrow$ & Timing $\uparrow$ & Content $\uparrow$ & Stability $\uparrow$ & \shortstack{AxE Score $\uparrow$} \\
\midrule
\multicolumn{6}{l}{\textit{Text-and-image-to-video}} \\
Kling 4.0 & \textbf{27.80} & 41.81 & \textbf{65.89} & \textbf{0.62} & \textbf{17.18} \\
Seedance 2.5 & 27.02 & \textbf{44.02} & 60.91 & \textbf{0.62} & 16.64 \\
Wan 3.0 & 24.70 & 42.90 & 57.07 & 0.60 & 14.70 \\
HappyHorse 1.1 & 21.68 & 38.21 & 56.59 & 0.55 & 11.96 \\
\cmidrule{1-6}
MiniMax H3 & \textbf{17.55} & 43.22 & \textbf{40.32} & \textbf{0.53} & \textbf{9.39} \\
LTX-2.5-Distilled & 17.14 & 43.73 & 38.28 & 0.52 & 8.84 \\
Cosmos3-Super & 15.63 & 40.82 & 38.16 & \textbf{0.53} & 8.34 \\
GigaWorld-0 & 15.15 & \textbf{47.79} & 31.57 & 0.47 & 7.18 \\
HunyuanVideo-1.5 & 14.15 & 38.20 & 36.95 & 0.46 & 6.44 \\
VPP2 & 13.30 & 39.36 & 33.65 & 0.43 & 5.67 \\
Wan2.2-TI2V-5B & 7.38 & 42.92 & 16.37 & 0.46 & 3.43 \\
Cosmos3-Nano & 8.17 & 34.54 & 23.37 & 0.37 & 3.03 \\
Cosmos-Predict2.5-2B & 7.08 & 44.46 & 15.30 & 0.37 & 2.63 \\
EchoWM & 5.51 & 26.91 & 20.48 & 0.31 & 1.70 \\
\midrule
\multicolumn{6}{l}{\textit{Camera-conditioned image-to-video}} \\
Wan2.2-Fun-Camera & \textbf{34.35} & 59.41 & 57.57 & 0.59 & \textbf{20.14} \\
SANA-WM & 28.09 & \textbf{64.47} & 43.10 & \textbf{0.67} & 18.94 \\
EchoWM Flash & 26.77 & 42.57 & \textbf{62.44} & 0.60 & 15.93 \\
Matrix-Game-2.0 & 24.52 & 59.35 & 41.32 & 0.64 & 15.75 \\
LingBot-World & 21.44 & 43.25 & 49.59 & 0.61 & 13.17 \\
Evoke & 12.82 & 26.89 & 45.49 & 0.46 & 5.91 \\
Evoke-Turbo & 11.91 & 26.37 & 44.44 & 0.45 & 5.41 \\
AlayaWorld & 9.13 & 35.65 & 25.62 & 0.37 & 3.38 \\
\bottomrule
\end{tabular*}
\endgroup
\end{table}

%% file: sections/6_conclusion.tex
\section{Conclusion}
\label{sec:conclusion}

We introduced \benchmarkname, which pairs each original condition with
variants that change the acting agent, the environment, or both along eight
factors, and WM-PerceptHarness, which scores a generated video by the events
it contains. On 1,103 recordings and 981 pairs, twenty models place events at
roughly the right times but rarely with the right content: no model exceeds a
Fidelity of 13.01. The scores also do not hold under a change. Stability is at most 0.55, so even the strongest model reaches only about half of its paired score on both conditions of a pair, and the highest AxE Score is 7.10. Applied to WBench, the same design gives the same picture for multi-turn tasks, with a Stability of at most 0.67. Fidelity follows human preference across models ($\rho = 0.90$), while no complementary metric exceeds 0.44. A deployed world model meets changed conditions routinely; current models do not yet carry what they get right in one condition over to a related one. These findings motivate evaluating event fidelity and robustness
jointly when assessing progress in video world models.